\documentclass[sigconf]{aamas} 
\usepackage{balance} % for balancing columns on the final page

\usepackage[utf8]{inputenc} % allow utf-8 input
\usepackage[T1]{fontenc}    % use 8-bit T1 fonts
\usepackage{hyperref}       % hyperlinks
\usepackage{url}            % simple URL typesetting
\usepackage{booktabs}       % professional-quality tables
\usepackage{amsfonts}       % blackboard math symbols
\usepackage{nicefrac}       % compact symbols for 1/2, etc.
\usepackage{microtype}      % microtypography
\usepackage{xcolor}         % colors
\usepackage{graphicx}
\usepackage{subfigure}
\usepackage{amsmath}
\usepackage{mathtools}
\usepackage{amsthm}
\usepackage{algorithmic}
\usepackage{algorithm}
\usepackage{verbatim}
\usepackage{tcolorbox}

\theoremstyle{plain}
\newtheorem{theorem}{Theorem}[section]
\newtheorem{proposition}[theorem]{Proposition}

\newtheorem{corollary}[theorem]{Corollary}
\theoremstyle{definition}
\newtheorem{definition}[theorem]{Definition}
\newtheorem{assumption}[theorem]{Assumption}
\theoremstyle{remark}

\newcommand{\awnn}{Hi-C } % The algorithm with no name
\newcommand{\awnnnospace}{Hi-C} 

\setcopyright{ifaamas}
\acmConference[AAMAS '24]{Proc.\@ of the 23rd International Conference
on Autonomous Agents and Multiagent Systems (AAMAS 2024)}{May 6 -- 10, 2024}
{Auckland, New Zealand}{N.~Alechina, V.~Dignum, M.~Dastani, J.S.~Sichman (eds.)}
\copyrightyear{2024}
\acmYear{2024}
\acmDOI{}
\acmPrice{}
\acmISBN{}

\acmSubmissionID{<<EasyChair submission id>>}

\title[Uncoupled Learning of Differential Stackelberg Equilibria with Commitments]{Uncoupled Learning of Differential Stackelberg Equilibria \\ with Commitments}

\author{Robert Loftin}
\authornote{Equal contribution.}
\affiliation{
  \institution{The University of Sheffield}
  \city{Sheffield}
  \country{United Kingdom}}
\email{r.loftin@sheffield.ac.uk}

\author{Mustafa Mert Çelikok}
\authornotemark[1]
\affiliation{
  \institution{Delft University of Technology}
  \city{Delft}
  \country{The Netherlands}}
\email{m.m.celikok@tudelft.nl}

\author{Herke van Hoof}
\affiliation{
  \institution{University of Amsterdam}
  \city{Amsterdam}
  \country{The Netherlands}}
\email{h.c.vanhoof@uva.nl}

\author{Samuel Kaski}
\affiliation{
  \institution{Aalto University}
  \city{Helsinki}
  \country{Finland} }
\affiliation{
  \institution{The University of Manchester}
  \city{Manchester}
  \country{United Kingdom}}
\email{samuel.kaski@aalto.fi}

\author{Frans A. Oliehoek}
\affiliation{
  \institution{Delft University of Technology}
  \city{Delft}
  \country{The Netherlands}}
\email{f.a.oliehoek@tudelft.nl}

\begin{abstract}
In multi-agent problems requiring a high degree of cooperation, success often depends on the ability of the agents to adapt to each other's behavior.  A natural solution concept in such settings is the Stackelberg equilibrium, in which the ``leader'' agent selects the strategy that maximizes its own payoff given that the ``follower'' agent will choose their best response to this strategy.  Recent work has extended this solution concept to two-player differentiable games, such as those arising from multi-agent deep reinforcement learning, in the form of the \textit{differential} Stackelberg equilibrium.  While this previous work has presented learning dynamics which converge to such equilibria, these dynamics are ``coupled'' in the sense that the learning updates for the leader's strategy require some information about the follower's payoff function.  As such, these methods cannot be applied to truly decentralised multi-agent settings, particularly ad hoc cooperation, where each agent only has access to its own payoff function.  In this work we present ``uncoupled'' learning dynamics based on zeroth-order gradient estimators, in which each agent's strategy update depends only on their observations of the other's behavior.  We analyze the convergence of these dynamics in general-sum games, and prove that they converge to differential Stackelberg equilibria under the same conditions as previous coupled methods.  Furthermore, we present an online mechanism by which symmetric learners can negotiate leader-follower roles.  We conclude with a discussion of the implications of our work for multi-agent reinforcement learning and ad hoc collaboration more generally.
\end{abstract}

\keywords{multi-agent reinforcement learning; ad hoc collaboration; ad hoc teamwork; learning dynamics; differentiable games; differential stackelberg equilibrium}
\newcommand{\BibTeX}{\rm B\kern-.05em{\sc i\kern-.025em b}\kern-.08em\TeX}

\makeatletter
\gdef\@copyrightpermission{
	\begin{minipage}{0.3\columnwidth}
		\href{https://creativecommons.org/licenses/by/4.0/}{\includegraphics[width=0.90\textwidth]{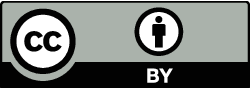}}
	\end{minipage}\hfill
	\begin{minipage}{0.7\columnwidth}
		\href{https://creativecommons.org/licenses/by/4.0/}{This work is licensed under a Creative Commons Attribution International 4.0 License.}
	\end{minipage}
	\vspace{5pt}
}
\makeatother

\begin{document}

%%% The following commands remove the headers in your paper. For final 
%%% papers, these will be inserted during the pagination process.

\pagestyle{fancy}
\fancyhead{}

%%% The next command prints the information defined in the preamble.

\maketitle

%%%%%%%%%%%%%%%%%%%%%%%%%%%%%%%%%%%%%%%%%%%%%%%%%%%%%%%%%%%%%%%%%%%%%%%%

\section{Introduction}
A central goal of multi-agent systems research has been to understand the long-term behavior of \textit{autonomous} learning agents that optimize their individual strategies through repeated interaction. The challenge here is that the agents do not have any direct control over each other, cannot see into to the "minds" of others, and must learn based solely on their observable behavior. In this work, we focus on this problem in collaborative settings, where agents benefit from cooperation, though they may not share the same payoff functions~\cite{mckee2020social,durugkar2021balancing}.  An important example of such a setting is \textit{ad hoc teamwork} \cite{stone2010ad,mirsky2022survey}, where agents that have never encountered one another before must learn to collaborate without any prior coordination. 
%Even though they benefit from cooperation, the agents need not have the same payoffs. For instance, they might have mixed-motives \cite{mckee2020social} or individual preferences \cite{durugkar2021balancing}. 
Such settings might arise when different companies create their own learning agents to interact with other agents or with humans. These agents will need to collaborate, bargain, and negotiate with each other, but would definitely prefer keeping their utilities and internal learning mechanisms private.

Theoretical analysis of multi-agent learning dynamics allows us to determine if and when autonomous learning agents will converge to a fixed joint strategy, characterize the strategies they are likely to converge to, and ultimately design improved learning algorithms (e.g. \cite{barfuss2022dynamical, 10.5555/3398761.3398822, 10.1007/s10458-023-09612-x}). Additionally, recent years have seen a surge of interest in the dynamics of multi-agent learning, driven by the recognition that many machine learning problems (such as actor-critic methods in RL~\cite{zheng2022stackelberg}) can be formulated as games with continuous, high-dimensional strategy spaces and differentiable payoff functions. Results on such \textit{differentiable} games have also found applications to multi-agent reinforcement learning~\cite{balduzzi2018mechanics}.

In this work we consider the problem of finding \textit{hierarchical} solutions in two-player, general-sum differentiable games. In the hierarchical model of play, the ``leader'' player selects their strategy first, after which the other player (the ``follower'') selects their best-response to this strategy. The natural solution concept for the hierarchical model is the \textit{Stackelberg equilibrium} (SE), in which the leader's strategy is optimal under the assumption that the follower will play their best response to whatever strategy the leader chooses. The hierarchical model is well suited to cooperative settings, where the leader can play their half of an optimal joint strategy knowing that the follower will respond appropriately. 

It has recently been argued that the Stackelberg equilibrium is also a more useful solution concept for differentiable games than the Nash equilibrium (NE), as the SE exists in games where the NE does not~\citep{jin2020local}. This fact has motivated the development of ``hierarchical'' gradient ascent methods for finding \emph{differential} Stackelberg equilibria (DSE), the local analogue of SE, in differentiable games.  In particular,~\citet{fiez2019stackelberg} have presented a hierarchical gradient update that is shown to converge to DSE in certain differentiable games. However, this \textit{coupled} learning update is designed with \emph{centralized training} in mind, and the leader's update requires knowledge of the follower's payoff function. Such coupled learning methods cannot be applied to independent learning settings, where the other agent's payoff function is unknown. The coupled hierarchical update also requires the Hessian of the follower's payoff function, and so may be computationally intractable in settings where second-order derivatives are expensive to estimate (such as reinforcement learning).  Finally, as with most existing literature, this learning update assumes that the roles of leader and follower are assigned beforehand (a form of prior coordination).

The main contribution of this work is a novel \textit{uncoupled} learning update called \textit{\underline{Hi}erarchical learning with \underline{C}ommitments} (\awnnnospace), which does not require the leader to have access to the follower's payoff function or learning algorithm. \awnn estimates the leader's gradient update by sampling strategies close to the leader's current strategy, and then committing to these ``perturbed'' strategies long enough that the follower has time to adapt to them.  As such, unlike previous coupled methods, the \awnn update is applicable to fully independent multi-agent learning settings such as ad hoc teamwork. As an added benefit, our method is a tractable alternative to coupled hierarchical updates for problems where estimating the higher-order derivatives of the payoff functions is possible but impractical. Our main theoretical results show that, under the same conditions as previous coupled methods, \awnn converges to a DSE for the leader as long as the follower's own strategy converges to its best response sufficiently fast. We mathematically specify what sufficiently fast means in this context, and as an illustrative example derive a commitment schedule for the case where the follower's payoff function is strongly concave.  Furthermore, we introduce a mechanism by which agents can ``negotiate'' the leader--follower role assignment in an online fashion.  This allows symmetric learners to negotiate their roles while simultaneously solving the underlying differentiable game.  To our knowledge, this is the first negotiation process that addresses the open question, presented in \citet{basar1973relative}, of determining roles online in hierarchical play.

%%%%%%%%%%%%%%%%%%%%%%%%%%%%%%%%%%%%%%%%%%%%%%%%%%%%%%%%%%%%%%%%%%%%%%%%

\section{Background}
\label{preliminaries}

% Differentiable games and notation

We consider the class of two-player, general-sum differentiable games.  Let $\mathcal{X} \subseteq \Re^{d_1}$ and $\mathcal{Y} \subseteq \Re^{d_2}$ be the strategy spaces for players 1 and 2 respectively.  In hierarchical play, we let player 1 be the leader and player 2 the follower, unless stated otherwise.  Let $f_{i} : \mathcal{X} \times \mathcal{Y} \mapsto \Re$ be the payoff function of player $i$, with $f_{i} \in C^2(\mathcal{X} \times \mathcal{Y}, \mathbb{R})$ for all $i \in \{1, 2\}$ (i.e. $f_{i}$ is twice continuously differentiable).  Let $\nabla_{x} f_{i}(x, y)$ and $\nabla_{y} f_{i}(x, y)$ denote the gradients of $f_{i}$ w.r.t. player 1 and player 2's strategies respectively.  We denote by $\nabla_{xy} f_{i}(x,y) = \nabla_y [\nabla_x f_{i}(x, y)]$ the Jacobian of the gradient $\nabla_x f_{i}(x,y)$ w.r.t. $y$, and define $\nabla_{yx} f_{i}(x,y)$, $\nabla_{xx} f_{i}(x,y)$, and $\nabla_{yy} f_{i}(x,y)$ similarly. Finally we let $\Vert \cdot \Vert$ denote the Euclidean norm throughout.

When analysing learning dynamics, the ``pure'' strategies $x$ and $y$ of the differentiable game can be thought of as representing the parameters of the (potentially stochastic) strategies followed by the agents in some underlying game.  For instance, any $N \times N$ matrix game can be described as a differentiable game by choosing $\mathcal{X}$ and $\mathcal{Y}$ to be the $N$-dimensional probability simplices. Then $f_i(x,y)$ would be the expected payoff for $i$ in the matrix game under the mixed strategy profile $\langle x,y \rangle$.  Deep reinforcement learning agents in a Markov game can be represented similarly, where $x$ and $y$ would be the parameters of neural networks representing stochastic policies.  The learning dynamics therefore capture how the players update their mixed strategies / stochastic policies over time.

\subsection{Simultaneous Gradient Ascent and Differential Nash Equilibria}

A straightforward approach to solving differentiable games is \textit{simultaneous gradient ascent} (SGA), where player $i$ performs gradient ascent on its own payoff function $f_i$, treating other players' strategies as fixed.  The two-player SGA updates are defined as
\begin{equation}
    x_{t+1} = x_t + \alpha_{1,t} \nabla_x f_1 (x_t, y_t) \qquad
    y_{t+1} = y_t + \alpha_{2,t} \nabla_y f_2 (x_t, y_t),    
\end{equation}
where the sequences $\{\alpha_{1,t}\}$ and $\{\alpha_{2,t}\}$ are learning rate schedules, which may differ between the players. SGA is often the default approach for problems described by two-player games (such as training GANs~\cite{gan_original}).  We can also view SGA as a model of ad hoc learning between independent agents. In the ad hoc setting, players are only aware of their own payoff functions, and the strategies other players follow at each \textit{stage} $t$ of the game.

%%%%%%%%%%%%%%%%%%%%%%%%%%%%%%%%%%%%
%%% Differential Nash Equilibria %%%
%%%%%%%%%%%%%%%%%%%%%%%%%%%%%%%%%%%%

However, as it is generally the case for gradient-based learners, we cannot expect SGA to always find global optima in the strategy space of either player. This motivates the development of  \textit{local} alternatives, including the differential Nash equilibrium (DNE).  
\begin{definition}[Differential Nash Equilibrium \cite{differential_nash}]
Let $\omega(x,y) = (\nabla_xf_1(x,y), \nabla_y f_2(x,y))$ be the individual gradients of the players' payoff functions at $(x,y)$. 
A strategy profile $(x^*, y^*) \in \mathcal{X} \times \mathcal{Y}$ is a differential Nash equilibrium if \textbf{(I)} $\omega(x^*, y^*) = 0$, \textbf{(II)} $\nabla_{xx}f_1(x^*,y^*)$ and $\nabla_{yy}f_2(x^*,y^*)$ are negative definite. 
\end{definition}

Previous work has shown that gradient-based learning dynamics such as SGA can converge to DNE in specific classes of games \cite{differential_nash, JMLR:v20:19-008}. However, the main issue with DNEs is that they fail to exist in some games, which constrains the class of games they are applicable to. For instance, Nash equilibria exist for convex costs (i.e. concave payoffs) on compact and convex strategy spaces, and DNE exists if these conditions, as described in \citet[Theorem~4.3 \& Chapter~4.9]{bacsar1998dynamic}, are met locally within the neighbourhoods DNE are defined \cite{fiez2020implicit}. An alternative local solution concept, discussed below, based on Stackelberg equilibria exists in more relaxed conditions, and is thus applicable to a wider class of games.

\subsection{Hierarchical Model and Differential Stackelberg Equilibria}

In the hierarchical model, the leader selects a strategy first, and the follower selects the best response to the leader's strategy. Thus, the natural solution concept in the hierarchical play is the Stackelberg equilibrium, in which the leader chooses a strategy that maximizes its payoff under the follower's best response.
\begin{definition}[Stackelberg Equilibrium (SE) \cite{simaan1973stackelberg}]
\label{def:stackelberg}
Let the set $\text{BR}(x) =\{ y \mid f_2(x,y) =\max_{y' \in \mathcal{Y}}  f_2(x, y') \}$ denote the follower's set of best-responses when the leader plays $x$.  A joint strategy $(x^*, y^*) \in \mathcal{X} \times \mathcal{Y}$ is a \textit{Stackelberg equilibrium} if $y^* \in \text{BR}(x^*)$ and
\begin{equation}
    \min_{y \in \text{BR}(x^*)} f_1(x^*, y) \geq \min_{y \in \text{BR}(x)} f_1(x, y)
\end{equation}
for all $x \in \mathcal{X}$.  Such an $x^*$ is a \textit{Stackelberg solution} for the leader.
\end{definition}
Note that for the SE to be well-defined, the follower must have a tie-breaking mechanism, and definition~\ref{def:stackelberg} assumes that the follower breaks ties so as to minimize the leader's payoffs. Therefore, a Stackelberg solution maximizes the leader's worst-case payoff assuming the follower will act rationally, and in zero-sum games the Stackelberg solution guarantees the leader will receive at least its security value. Recent work~\citep{jin2020local, fiez2019stackelberg} has shown that the hierarchical model can be applied to differentiable games as well. While a differentiable game may possess no Nash equilibria, a Stackelberg equilibrium will always exist so long as the strategy spaces $\mathcal{X}$ and $\mathcal{Y}$ are compact~\citep[Theorem~4.8 \& Chapter~4.9]{bacsar1998dynamic}. Algorithms based on the hierarchical model have proven successful in training generative adversarial networks~\cite{fiez2020implicit, metz2017unrolled} and actor--critic methods~\cite{zheng2022stackelberg}.  

Definition~\ref{def:stackelberg} assumes that both the leader and the follower have found \textit{global} optima in their respective strategy spaces. %For non-concave payoff functions, the best an individual player can hope to find with gradient ascent is a \textit{local} optimum of its individual objective (even if the other player's strategy remained fixed). 
As gradient-based learners cannot guarantee convergence to a global optimum in general, applying the hierarchical model to differentiable games requires a local version of the SE referred to as the \emph{differential Stackelberg equilibrium} (DSE)~\cite{fiez2020implicit}.  The definition of DSE starts with the following observation. Given a point $(x^*, y^*) \in \mathcal{X} \times \mathcal{Y}$ such that $\nabla_y f_2(x^*, y^*) = 0$ and $det(\nabla_{yy} [f_2(x^*, y^*)]) \neq 0$, there exists a continuously differentiable local best-response function $r : \mathcal{U}_1 \mapsto \mathcal{Y}$ for the follower, where $\mathcal{U}_1 \subset \mathcal{X}$ is a neighbourhood of $(x^*, y^*)$. Under this notation, the leader's objective function becomes $f_1(x, r(x))$, and so a local optimum $x^*$ for the leader will satisfy $\nabla_x [ f_1(x, r(x)) ] = 0$, where $\nabla_x [ f_1(x, r(x)) ] = \nabla_x f_1(x, r(x)) + [\nabla_y f_1(x, r(x)]^{\top} \nabla_x r(x)$. Then, a DSE is defined as follows.

\begin{definition}[Differential Stackelberg Equilibrium \cite{fiez2020implicit}]
\label{def:differential_stackelberg}
A strategy profile $(x^*, y^*) \in \mathcal{X} \times \mathcal{Y}$ with $r(x^*) = y^*$, 
is a differential Stackelberg equilibrium (DSE) if: \textbf{(I)} $\nabla_x [f_1(x^*, r(x^*))] = 0$ and $\nabla_y [f_2(x^*, y^*)] = 0$, and \textbf{(II)} $\nabla_{xx} [f_1(x^*, r(x^*))]$ and $\nabla_{yy} [f_2(x^*, y^*)]$ are negative definite.  Furthermore, any $x^*$ satisfying these conditions is a differential Stackelberg solution (DSS) for the leader.
\end{definition}

Intuitively, the condition \textbf{(II)} ensures that $x^*$ and $y^*$ are local maxima of the player's individual objectives, rather than minima or saddle points.  Note that conditions \textbf{(I)} and \textbf{(II)} do not imply that $\nabla_x f_1(x^*, y^*) = 0$, and so DSE may not always be stable under gradient ascent on $f_1$. In fully-cooperative games, all agents have the same reward function (i.e. $f_1 = f_2$). In that case, we have the following proposition, which states that learning the DSE instead of DNE does not break results for the fully-cooperative case.

\begin{proposition}[Fully-cooperative Multi-agent RL and DSE]
\label{prop:equivalence}
    Differential Stackelberg equilibria and differential Nash equilibria are equivalent in fully-cooperative games where $f_1 = f_2$.
\end{proposition}  %%% REMOVED EXPLICIT MENTION OF MULTI-AGENT RL, AS THIS ISN'T PRECISELY DEFINED

\subsection{Hierarchical Gradient Update}

While a natural approach to finding DSE is to perform gradient ascent on the leader's objective function $f_1(x, r(x))$, we will generally not have a closed-form expression for $r(x)$.  Fortunately, for a joint strategy $(x, y) \in \mathcal{X} \times \mathcal{Y}$ for which $y = r(x)$ and $\nabla_{yy} f_2 (x,y)$ is non-singular, the implicit function theorem provides a closed-form expression for $\nabla_x r(x)$ as a function of $x$ and $y$~\cite{fiez2019stackelberg}.  When $y = r(x)$, we have $\nabla_x r(x) = -\left( \nabla_{yy} f_2(x, y) \right)^{-1} \nabla_{xy} f_2(x, y)$.  The gradient of the leader's objective then becomes
\begin{align}
    \nabla_x [f_1(x, r(x))] = \nabla_x &f_1(x, y) \nonumber \\   
    &-\nabla_y f_1(x, y)^{\top}\left( \nabla_{yy} f_2(x, y) \right)^{-1} \nabla_{xy} f_2(x, y) \nonumber \\
    &= D(x,y) \label{eqn:hierarchical_gradient}
\end{align}
Evaluating~\ref{eqn:hierarchical_gradient} requires knowing the value of $y = r(x)$.  One way to compute the leader's gradient update is then to optimize $y$ via gradient ascent on $f_2$ while keeping $x$ constant, and allowing $y$ to converge to $r(x)$ before performing each gradient step for the leader's strategy.  \citet{fiez2019stackelberg} present a more practical, two-timescale algorithm in which the leader and follower strategies are updated simultaneously, with the follower using a faster learning rate than the leader.  When we may only have noisy estimates of the gradients, the two-timescale hierarchical gradient updates become:
\begin{align}
    &x_{t+1} = x_{t} + \alpha_{1,t}(D (x_t, y_t) + w_{1,t}) \nonumber \\ 
    &y_{t+1} = y_{t} + \alpha_{2,t}(\nabla_y f_2(x_t, y_t) + w_{2,t}) 
    \label{eqn:leader_follower_dynamics},
\end{align}
where $\{ w_{1,t} \}$ and $\{ w_{2,t}\}$ are independent zero-mean noise sequences, and the leader's update $D(x,y)$ is defined as in Equation~\ref{eqn:hierarchical_gradient}.  To achieve time-scale separation, the learning rate schedules are chosen so that $\alpha_{2,t} \gg \alpha_{1,t}$, which allows the follower's strategy to ``track'' its best response to the leader's current strategy.  If the limit condition $\lim_{t \rightarrow \infty}{\frac{\alpha_{1,t}}{\alpha_{2,t}}} = 0$ holds, then results on two-timescale stochastic approximation (see~\cite[Chapter 6.1]{borkar2009stochastic}) can be used to analyze the convergence properties of~\ref{eqn:leader_follower_dynamics}.  

\subsection{Limitations of Coupled Learning}

We can see that the explicit form of the leader's update in Equation~\ref{eqn:hierarchical_gradient} depends on the Hessian $\nabla_{yy} f_2$ of the follower's payoff function, which implies that the leader must know the structure of $f_2$.  This assumption does not hold in ad hoc cooperation, where the leader only has access to the follower's observable behavior. Even in centralized settings where the leader can estimate the gradient and Hessian of $f_2$ directly, this estimation can be expensive and suffer from high variance.  This is particularly true in settings such as reinforcement learning, where gradients (and Hessians) must be estimated through Monte-Carlo simulations.  Other learning updates such as LOLA also depend on estimates of the follower Hessian~\citep{foerster2018lola}, and so suffer form the same limitations.  In this section, we remove the limitation of coupled learning by introducing a learning algorithm that estimates $\nabla_x [f_1(x, r(x))]$ from the follower's behavior alone, while maintaining similar convergence guarantees to the two-timescale hierarchical gradient update.

%%%%%%%%%%%%%%%%%%%%%%%%%%%%%%%%%%%%%%%%%%%%%%%%%%%%%%%%%%%%%%%%%%%%%%%%
\section{Uncoupled Learning with Commitments}
\label{hi_c}

\begin{algorithm}[h]
\begin{algorithmic}[1]
    \STATE \textbf{Inputs:} Step-sizes $\{ \alpha_n \}_{n \geq 0}$, perturbation schedule $\{ \delta_n \}_{n \geq 0}$, commitment schedule $\{ k_n \}_{n \geq 0}$.
    \STATE \textbf{Initialize:} sample $x_0$ from $\mathcal{X}$
    \FOR{step $n = 0,1,\ldots$}
        \STATE sample $\Delta_n$ uniformly from $\{ -1, 1 \}^{d_1}$.
        \STATE $\tilde{x}_n \leftarrow x_n + \delta_n \Delta_n$
        \FOR{stage $t = t(n), \ldots, t(n) + k_n - 1$}
            \STATE play $\tilde{x}_n$.
            \STATE observe $\tilde{y}_{n} \leftarrow y_t$.
        \ENDFOR
        \FOR{dimension $i = 1, \ldots, d_1$}
            \STATE $x^{i}_{n+1} = x^{i}_{n} + \frac{\alpha_n}{\delta_n \Delta^{i}_n} [f_1(\tilde{x}_n, \tilde{y}_n) + w_n]$
        \ENDFOR
    \ENDFOR
\end{algorithmic}
\caption{The \awnn learning algorithm, with follower strategies $y_t$ chosen arbitrarily, and with $w_n$ being some zero-mean noise.  $t(n) = \sum^{n-1}_{m=0} k_m$ is the stage at which interval $n$ started.}
\label{alg:hi_c}
\end{algorithm}

From the leader's perspective, the problem of finding a differential Stackelberg equilibrium is simply that of finding a local maximum of $f_1 (x, r(x))$, where $r(x)$ is the follower's best response when the leader chooses $x$ as their strategy.  The challenge in the \textit{uncoupled} setting is that the leader cannot evaluate $\nabla_x [f_1 (x, r(x))]$ directly, since it cannot evaluate the Jacobian $\nabla_x r(x)$ as it does not have access to the follower's payoff function $f_2$ on which $r(x)$ depends.  The leader can, however, estimate the value of $r(x)$ (and therefore $f_1(x, r(x))$) by simply observing the follower's response when it plays strategy $x$.  A natural approach then is to replace gradient ascent with a gradient-free learning rule that only requires an unbiased estimate of $f_1(x, r(x))$, and not of $\nabla_x [f_1 (x, r(x))]$.

We first consider the hypothetical case where the leader has access to an \textit{oracle} for $r(x)$.  This oracle allows the leader to evaluate $f_1(x, r(x))$ for any $x \in \mathcal{X}$.  We can then apply \textit{simultaneous perturbation stochastic approximation} (SPSA)~\cite{spall199spsa} to approximate gradient ascent on $f_1(x, r(x))$.  Specifically, we will derive \awnn from the one-sample form of SPSA~\cite{spall1997one}. For all $n \geq 0$, let $\Delta_n$ be independently and uniformly sampled from $\{ -1, 1 \}^{d_1}$, and let $\{ \delta_n \}_{n \geq 0}$ be a decreasing \textit{perturbation schedule}.  Let $\{ w_n \}_{n \geq 0}$ be a sequence of i.i.d. variables (with zero-mean and uniformly bounded variance) representing noise in the evaluation of $f_1$.  The element-wise one-sample SPSA update is then
\begin{equation}
\label{eqn:spsa}
    x^{i}_{n+1} = x^{i}_{n} + \alpha_n \frac{f_1(x_n + \delta_n \Delta_n, r(x_n + \delta_n \Delta_n)) + w_n}{\delta_n \Delta^{i}_n}     
\end{equation}
for all $i \in [1, d_1]$.  SPSA estimates the direction of the gradient by sampling points near the current strategy $x_n$.  Going forward, let $\tilde{x}_n = x_n + \delta_n \Delta_n$ denote the ``perturbed'' strategy evaluated at step $n$.  The noise terms $w_t$ account for settings the leader can only observe an unbiased estimator of $f_1$ (e.g., a single policy roll-out).

\paragraph{Estimating $r(\tilde{x}_n)$.} In truly uncoupled settings the leader has no way of directly computing $r(x_n)$.  What the leader can do is observe the strategies played by the follower, which presumably updates its own strategy so as to maximize its payoff under $f_2$.  This suggests an asynchronous, two-timescale learning process, in which the leader \textit{commits} to playing the perturbed strategy $\tilde{x}_n$ for some $k_n$ stages before updating $x_n$.  For sufficiently large $k_n$ we should hope that after $k_n$ stages the follower's strategy will have approximately converged to its best-response $r(\tilde{x}_n)$.

Under the \awnn learning update (Algorithm~\ref{alg:hi_c}), at each interval $n \geq 0$ the leader samples a perturbed strategy $\tilde{x}_n$, and then plays this strategy for the next $k_n$ stages. After $k_n$ stages, the leader updates its strategy element-wise as
\begin{equation}
\label{eqn:hi_c}
    x^{i}_{n+1} = x^{i}_{n} + \alpha_n \frac{f_1(\tilde{x}_n, \tilde{y}_{n}) + w_n}{\delta_n \Delta^{i}_n}     
\end{equation}
where the follower's final strategy within interval $n$, denoted by $\tilde{y}_{n}$, is used as an estimate of $r(\tilde{x}_n)$. This gradient estimator has bounded variance, since $f_1$ is bounded on $X \times Y$ and $w_n$ is i.i.d. noise with bounded variance.  To reduce variance we can optionally use $f_1(x_t, \tilde{y}_{n-1})$ as a baseline value, as $\tilde{y}_{n-1}$ is independent of $\Delta_n$.

\subsection{Convergence Analysis}
\label{convergence}

In this section we make no assumptions about the specifics of the follower's learning update or their payoff function, and instead prove convergence of the leader's strategy under a generic assumption about the convergence rate of the ``tracking error'' $\Vert \tilde{y}_n - r(\tilde{x}_n)\Vert$ between the follower's strategy and its best-response (asm.  \ref{asm:tracking_error}). In general, this assumption can be satisfied by choosing a long enough commitment schedule for the leader. As an illustrative example, in Section~\ref{commitments} we derive a commitment schedule $\{ k_n \}_{n \leq 0}$ for followers with strongly concave payoff functions that ensures the tracking error will decrease fast enough to satisfy this assumption.

In \awnnnospace, the follower is assumed to update their strategy at every stage $t$, while the leader only performs an update after $k_n$ stages. Let $t(n) = \sum^{n-1}_{m=0} k_m$ be the stage at which the leader begins its $n$th commitment interval, and let $n(t) = \max\{ n : t(n) \leq t\}$ be the current interval at stage $t$.  We let $x_n$ ($n \geq 0$) denote the leader's \textit{mean} strategy after $n$ updates, or $t(n)$ stages, and let $y_t$ denote the strategy the follower played at stage $t$.  We then have $\tilde{y}_{n + 1} = y_{(t(n) + k_n - 1)}$, the last strategy the follower played during the $n$th commitment interval.  We prove convergence of \awnn under  assumptions that are standard in the analysis of previous work from simultaneous perturbation methods~\cite[Chapter 5]{bhatnagar2012stochastic} and differential Stackelberg equilibria (asm. \ref{asm:uniqueness}, \ref{asm:boundedness}, 
\ref{asm:smoothness}, and \ref{asm:step_size}). 

\begin{assumption}
\label{asm:uniqueness}
There exists a unique best-response function $r : \mathcal{X} \mapsto \mathcal{Y}$ that maps leader strategies to follower's best-responses. Furthermore, $r$ is $L_r$-Lipschitz and $K_r$-smooth.
\end{assumption}

Note that the assumption \ref{asm:uniqueness} does not restrict the follower's payoff function to have a unique optimizer for a given $x$, but simply requires that the follower breaks ties in an arbitrary yet fixed way. This is a common assumption in hierarchical play, and it is needed for the SE to be well-defined. The leader also does not make any assumptions on how the follower breaks ties; the tie-breaking mechanism is abstracted away into the follower's best-response function $r$, which is estimated from observed behaviour.

\begin{assumption}
\label{asm:boundedness}
$x_n$ and $y_t$ are bounded almost surely:
\begin{equation}
    \sup_{n \geq 0} \Vert x_n \Vert < \infty \quad \text{and} \quad \sup_{t \geq 0} \Vert y_t \Vert < \infty \quad \text{a.s.}
\end{equation}
\end{assumption}

This immediately implies that $\tilde{x}_n$ and $\tilde{y}_n$ are bounded a.s., and because $r$ is Lipschitz, it implies $r(\tilde{x}_n)$ is bounded a.s. as well.  In practice, the assumption that the strategies remain bounded can be enforced by choosing $\mathcal{X}$ and $\mathcal{Y}$ to be bounded, and projecting the strategies back to $\mathcal{X}$ and $\mathcal{Y}$ whenever necessary.

\begin{assumption}
\label{asm:smoothness}
The leader's payoff function $f_1(x, y)$ is $L_1$-Lipschitz, and $K_1$-smooth in both of its arguments.
\end{assumption}

Assumption \ref{asm:smoothness} implies that $\Vert \nabla_y f_1(x,y) \Vert \leq L_1$ for all $(x, y) \in \mathcal{X} \times \mathcal{Y}$.  Combined with Assumption~\ref{asm:uniqueness}, it also implies that the leader's hierarchical objective function $g(x) = f_1(x, r(x))$ is also Lipschitz and smooth.
\begin{assumption}
\label{asm:step_size}
The step-size schedule $\{ \alpha_n \}_{n \geq 0}$ and perturbation schedule $\{ \delta_n \}_{n \geq 0}$ satisfy
\begin{align}
    \lim_{n \rightarrow \infty} \alpha_n = 0, \quad \lim_{n \rightarrow \infty} \delta_n = 0, \quad
    \sum^{\infty}_{n=0} \alpha_n = \infty, \quad \sum^{\infty}_{n=0} \frac{\alpha^2_n}{\delta^2_n} < \infty.
\end{align}
\end{assumption}

The decreasing magnitude $\delta_n$ of the perturbations means that eventually even small errors in the approximation of $r(\tilde{x}_n)$ could lead to large errors in the estimate of the gradient. Therefore, other than the standard assumptions listed above, the following generic assumption on the rate of convergence of the follower's tracking error must be satisfied via an appropriate commitment schedule.

\begin{assumption}
\label{asm:tracking_error}
Define $\varepsilon_n = \Vert \tilde{y}_n - r(\tilde{x}_n) \Vert$, for $n \geq 0$.  The commitment and perturbation schedules $\{ k_n \}_{n \geq 0}$ and $\{ \delta_n \}_{n \geq 0}$ satisfy
\begin{equation}
    \lim_{n \rightarrow \infty} \frac{\varepsilon_n}{\delta_n} = 0 \quad \text{and} \quad \sup_{n \geq 0} \frac{\varepsilon_n}{\delta_n} < \infty \quad \text{a.s.}
\end{equation}
\end{assumption}

Under Assumption~\ref{asm:tracking_error}, the error introduced by using $\tilde{y}_n$ rather than $r(\tilde{x}_n)$ is bounded and $o(1)$ almost surely, and so becomes negligible asymptotically.  To see this, we rewrite Equation~\ref{eqn:hi_c} as
\begin{equation}
    x^{i}_{n+1} = x^{i}_n + \alpha_n \left ( \frac{f_1 (\tilde{x}_n, r(\tilde{x}_n)) + w_t}{\delta_n \Delta^{i}_n} + \eta^{i}_n \right)
\end{equation}
where
\begin{align}
    \eta^{i}_n = \frac{f_1(\tilde{x}_n, \tilde{y}_{n+1}) - f_1(\tilde{x}_n, r(\tilde{x}_n))}{\delta_n \Delta^{i}_n} \leq \frac{L_1 \Vert \tilde{y}_n - r(\tilde{x}_n) \Vert}{\delta_n \Delta^{i}_n}
\end{align}
% We can make this much simple using the Lipschitz assumption
since $f_1$ is $L_1$-Lipschitz by Assumption~\ref{asm:smoothness}.  We then have
\begin{equation}
    \sup_{n \geq 0} \vert \eta^{i}_n \vert \leq \sup_{n \geq 0} \frac{L_1 \Vert \tilde{y}_n - r( \tilde{x}_n ) \Vert}{\delta_n} = \sup_{n \geq 0} L_1 \frac{\varepsilon_n}{\delta_n} < \infty \quad \text{a.s.} \\
\end{equation}
where the final inequality comes from the fact that $\sup_{n \geq 0}\frac{\varepsilon_n}{\delta_n}$ is bounded almost surely by Assumption~\ref{asm:tracking_error}.  We can also see that $\lim_{n \rightarrow \infty} \vert \eta^{i}_n \vert = 0$ almost surely, as $\frac{\varepsilon_n}{\delta_n} \rightarrow 0$ a.s.. We are now ready to state our main convergence result:

\begin{theorem}
\label{thm:convergence}
Let $H \subseteq \mathcal{X}$ be the set $\{ x \in  \mathcal{X} : \nabla_x [f_1 (x, r(x))] = 0\}$.  Assume $H \neq \emptyset$, and that Assumptions~\ref{asm:uniqueness}--\ref{asm:tracking_error} are satisfied under the \awnn update (Algorithm~\ref{alg:hi_c}).  Then the leader's strategy $x_n$ will converge to $H$ almost surely as $n \rightarrow \infty$. 
\end{theorem}

%\footnote{A more detailed proof is given in appendix A.}
\textit{Proof sketch:} This result follows  from~\citet[Theorem~5.2]{bhatnagar2012stochastic} by noting that the \awnn update in Equation~\ref{eqn:hi_c} is equivalent to the single measurement SPSA update (Equation~\ref{eqn:spsa}) save for the bounded, $o(1)$ error term $\eta^{i}_n$, which becomes negligible asymptotically (see~\cite[Chapter~2]{borkar2009stochastic}. This result shows that the \awnn update converges to a relatively small subset of $\mathcal{X}$ that, if they exist, contains the differential Stackelberg solutions. 
With further assumptions on $H$, we can show that $x_n$ converges to a DSS almost surely.

\begin{corollary}
\label{cor:dse_convergence}
Additionally, assume that $H$ consists only of isolated, asymptotically stable equilibria of the ODE $\dot{x}(t) = \nabla_x [f(x(t), r(x(t)))]$.  Then, under the \awnn update, $x_n$ will converge to a differential Stackelberg solution of the game $(f_1, f_2)$ almost surely as $n \rightarrow \infty$.
\end{corollary}

This follows from the fact that if $x \in H$ is an asymptotically stable equilibrium of $\dot{x}(t) = \nabla_x [f(x(t), r(x(t)))]$, then the Hessian $\nabla_{xx} [f(x(t), r(x(t)))]$ must be negative definite.  Combined with $\nabla_{x} [f(x(t), r(x(t)))] = 0$, this satisfies the requirements of Definition~\ref{def:differential_stackelberg}.  At first it may seem contradictory that we can prove convergence to a DSS when these are not guaranteed to exist.  The conditions under which Corollary~\ref{cor:dse_convergence} holds true, however, are precisely those conditions under which a DSS does exist, that is, when $f_1(x, r(x))$ has a strict local minimum in $\mathcal{X}$. 

Note that these results make no direct assumptions about the follower's payoff function or learning update.  Indeed, if we relax the second part of Assumption~\ref{asm:uniqueness} they could be satisfied for finite $\mathcal{Y}$ and discontinuous $r(x)$.  We simply require that for every $x \in \mathcal{X}$ the follower's strategy will converge to some unique fixed point $r(x)$ at a sufficiently fast rate \textit{relative} to the leader's commitment schedule.  In the next section we will consider some specific scenarios in which this requirement is satisfied, and how we can select a suitable commitment schedule given some information about the follower.

\subsection{Choosing the Commitment Schedule}
\label{commitments}

In order to derive specific commitment schedules that provably satisfy Assumption~\ref{asm:tracking_error}, we need finite-time convergence rate guarantees for the follower's strategy. It is important to note that the Hi-C algorithm itself does not require the knowledge of the convergence rate. The leader can always choose the commitment schedule with respect to the worst known upper-bounds of first-order optimisation methods, assuming the slowest rate for the follower. However, when more is known about the follower's rate of convergence, we can use the rate to derive better commitment schedules. To illustrate how to derive $k_n$ from known rates, we will consider one such well-studied case where the follower's objective function is strongly concave.
% In appendix B, we present a more general case of known convergence rates for $f_2$ (semi-strong concavity). 
Throughout this section we will make additional assumptions on the payoff functions $f_2$, and the best-response function $r$. These assumptions are needed only for the results within section \ref{commitments}.

\begin{assumption}
\label{asm:convexity}
$\forall x \in \mathcal{X}$, $f_2(x,y)$ is $K_2$-smooth and $\mu$-strongly concave w.r.t. $y$.
\end{assumption}

Under the assumption \ref{asm:convexity}, deterministic gradient ascent on $f_2$ with a fixed step-size schedule $\beta_t = \beta$ is sufficient for the follower's strategy to converge to its best-response.

\begin{proposition}[{\citet[Chapter 2]{nesterov2018convex}}]
\label{prop:convex_convergence}
Let the follower update its strategy using deterministic gradient ascent with a fixed step-size $\beta \in (0,\frac{1}{K_2}]$, such that
\begin{equation}
    y_{t+1} = y_t + \beta \nabla_y f_2(\tilde{x}_{n(t)}, y_t)
\end{equation}
then for any stage $t \geq 0$, and any $k \in [1, k_{n(t)}]$, we have
\begin{equation}
    \Vert y_{t+k} - r(\tilde{x}_{n(t)}) \Vert \leq (1 - \beta \mu)^{\frac{k}{2}} \Vert y_t - r(\tilde{x}_{n(t)}) \Vert.
\end{equation}
\end{proposition}

Imagine we are given step-size and perturbation schedules $\{ \alpha_n \}_{n \geq 0}$ and $\{ \delta_n \}_{n \geq 0}$ satisfying Assumption~\ref{asm:step_size}.  To find a suitable commitment schedule, we choose an arbitrary sequence $\{ \xi_n \}_{n \geq 0}$ such that $\lim_{n \rightarrow \infty} \xi_n = 0$ and $\sup_{n} \xi_n < \infty$.  We then need a commitment schedule $\{ k_n \}_{n \geq 0}$ such that:
\begin{equation}
    \frac{1}{\delta_n} \Vert \tilde{y}_n - r(\tilde{x}_n) \Vert \leq \xi_n
\end{equation}
for all $n \geq 0$.  To apply Proposition~\ref{prop:convex_convergence}, we need to be able to bound $\Vert y_t - r(\tilde{x}_{n(t)}) \Vert$ for all $t \geq 0$.  Previously we simply required that the strategies be bounded almost surely (Assumption~\ref{asm:boundedness}).  Now will will assume that this bound is deterministic, and known in advance.

\begin{assumption}
    \label{asm:sure_boundedness}
    There exists a deterministic constant $B < \infty$ such that $\sup_{t \geq 0} \Vert y_t \Vert < \frac{B}{2}$ and $\sup_{n \geq 0} \Vert r(\tilde{x}_n) \Vert <  \frac{B}{2}$ almost surely.
\end{assumption}

This always holds if $y_t$ is constrained to a bounded set $\mathcal{Y}$. We then have $\sup_{t \geq 0} \Vert y_t - r(\tilde{x}_n(t)) \Vert \leq B$ almost surely.  Then, under Assumptions~\ref{asm:convexity} and~\ref{asm:sure_boundedness}, using the proposition \ref{prop:convex_convergence}, we have that
\begin{equation}
    \frac{1}{\delta_n} \Vert \tilde{y}_n - r(\tilde{x}_n) \Vert \leq \frac{1}{\delta_n}(1 - \beta\mu)^{\frac{k_n}{2}} B.
\end{equation}
Upper-bounding this by $\xi_n$, we have
\begin{align}
    \frac{1}{\delta_n}(1 - \beta\mu)^{\frac{k_n}{2}} B &\leq \xi_n \\
    % \frac{k_n}{2} \ln{(1 - \beta\mu)} &\leq \ln{\frac{\delta_n \xi_n}{B}} \\
    2\frac{\ln{\delta_n \xi_n} - \ln{B}}{\ln{(1-\beta\mu)}} &\leq k_n. 
\end{align}
Setting $\xi_n \!\!=\!\! \frac{1}{n^p}$ for $p \!>\! 0$, we obtain the following convergence result:

\begin{corollary}
\label{cor:convex_convergence}
For the leader's perturbation schedule $\{ \delta_n \}_{n \geq 0}$, define the commitment times as
\begin{equation}
    \label{eqn:commitment}
    k_n = \left\lceil 2\frac{\ln \delta_n -\ln B - p \ln n}{\ln (1 - \beta\mu)} \right\rceil
\end{equation}
for $p > 0$ and $\beta \in (0, \frac{1}{K_2}]$.  Under Assumptions~\ref{asm:uniqueness} through~\ref{asm:step_size},~\ref{asm:convexity} and~\ref{asm:sure_boundedness}, if the follower updates their strategy using deterministic gradient ascent with step-size $\beta$, then the leader strategies $x_n$ computed by \awnn converge to $H$ almost surely as $n \rightarrow \infty$. 
\end{corollary}

The specified $k_n$ gives us $\frac{1}{\delta_n} \Vert \tilde{y}_n - r(\tilde{x}_n) \Vert \leq \frac{1}{n^p}$, and so $\{ k_n \}_{n \geq 0}$ satisfies Assumption~\ref{asm:tracking_error}.  The result then follows immediately from Theorem~\ref{thm:convergence}.  For $\delta_n = \frac{\bar{\delta}}{n^q}$, this yields $k_n = O(\ln n)$.

The assumption that the follower's payoffs are strongly concave
%(or semi-strongly in appendix B) 
makes intuitive sense in this setting. The only information the leader can obtain about the follower's asymptotic best-responses is through their finite time adaptation to the leader's current strategy.  If, over some subset $S \subset \mathcal{Y}$ containing $r(x)$, the curvature of $f_2$ is allowed to be arbitrarily small, then once the follower's strategy reaches $S$ it may converge to $r(x)$ arbitrarily slowly.  From the leader's perspective, it would appear that the follower has nearly settled on a best-response, such that the leader may over- or under-estimate the value of their current strategy.  It is therefore reasonable to require that the leader have some information about how fast the follower's strategy should converge.  In Corollary~\ref{cor:convex_convergence}, the necessary commitment schedule depends on $\beta\mu$, which determines the follower's convergence rate.

\subsection{Numerical Experiments}
\label{experiments}

\begin{figure}
    \centering
    \includegraphics[scale=0.5]{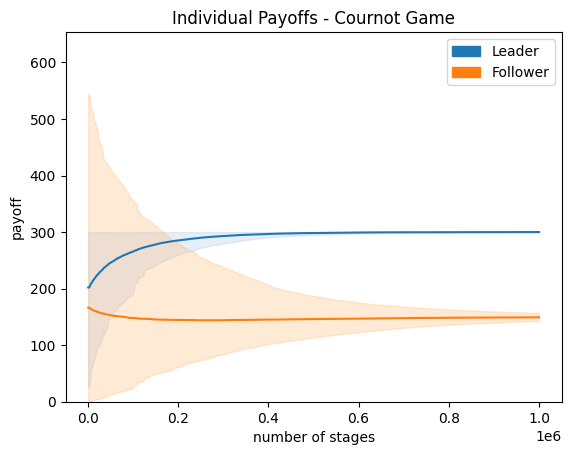}
    \caption{\awnn paired with gradient ascent in the Cournot duopoly.  Averaged over 32 runs (shaded regions show ranges).}
    \label{fig:cournot}
\end{figure}

We demonstrate the convergence behavior of \awnn in a simple differentiable game corresponding to the Cournot duopoly model~\cite{shapiro1989oligopoly} with linear prices and costs.  Here $x,y \in \Re$ are the quantities of some good produced by each player, with payoffs defined as
\begin{align}
    f_1(x, y) &= x [50 - (x + y)] - x \\
    f_2(x, y) &= y [50 - (x + y)] - y
\end{align}
The payoff each player receives depends on the price per unit, which is a strictly decreasing function of the total quantity produced.  Note that the unique Stackelberg equilibrium of this game (as well as the unique DSE) is $x=24.5$, $y=12.25$, such that $f_1(x,y) \approx 300$ and $f_2(x, y) \approx 150$.  Figure~\ref{fig:cournot} shows that \awnn converges to this solution when paired with a gradient ascent learner in the Cournot game.  Initially the gradient ascent learner (the follower) maximizes its payoff under the assumption that the leader's strategy is fixed.  Over time \awnn (the leader) increases its production quantity, knowing that the follower will reduce its own production to maintain its profits.  In these experiments \awnn uses a learning rate schedule of $\alpha_n = .001 n^{-.1}$ and a perturbation schedule of $\delta_n = n^{-.6}$, with the corresponding commitment schedule computed by Equation~\ref{eqn:commitment}. The gradient ascent learner uses a fixed learning rate of $0.1$.\footnote{Code available at: \url{https://github.com/rtloftin/Hi-C/tree/aamas2024}}

%%%%%%%%%%%%%%%%%%%%%%%%%%%%%%%%%%%%%%%%%%%%%%%%%%%%%%%%%%%%%%%%%%%%%%%%

\section{Role Negotiation}

\begin{algorithm}[h]
\begin{algorithmic}[1]
    \STATE \textbf{Inputs:} Step-sizes $\{ \beta_n \}_{n \geq 0}$, perturbation schedule $\{ \kappa_n \}_{n \geq 0}$, commitment schedule $\{ \tau_n \}_{n \geq 0}$.
    \STATE \textbf{Initialize:} $w_0 \leftarrow 0$
    \FOR{step $n = 0,1,\ldots$}
        \STATE sample $\Delta$ uniformly from $\{ -1, 1 \}$.
        \STATE $\tilde{w}_n \leftarrow w_n + \kappa_n \Delta_n$.
        \STATE sample $z_n$ uniformly from $[0, 1]$.
        \IF{$z_n < \sin^2 (\tilde{w}_n)$}
            \STATE follow the \awnn update (Algorithm~\ref{alg:hi_c}) on $f_1$ for $\tau_n$ stages
        \ELSE
            \STATE follow the gradient ascent update on $f_1$ for $\tau_n$ stages
        \ENDIF
        \STATE collect observed joint strategies $\{ \bar{x}^{n}_i \}$ and $\{ \bar{y}^{n}_i \}$
        \STATE $a_n \leftarrow \frac{1}{\tau_n} \sum^{\tau_n}_{i=1} f_1 (x^{n}_i, y^{n}_i)$
        \STATE $w_{n+1} = w_{n} + \frac{\beta_n a_n}{\kappa_n \Delta_n}$
    \ENDFOR
\end{algorithmic}
\caption{The meta-learning update for role negotiation.  Maintains a ``meta-strategy'' parameterized by $w_n$, and samples its role from its current strategy at each meta-learning interval.}
\label{alg:meta_learning}
\end{algorithm}

So far we have assumed that the leader and follower roles are assigned by some external process.  In this section we will briefly explore ways in which uncoupled learners might ``negotiate'' who will lead and who will follow during the training process itself.  In principle, two independent agents could use a variety of protocols (e.g., a coin toss) to agree upon their respective roles before training begins.  Here, however, we consider whether roles can themselves be \textit{learned} in a way that maximizes each player's individual payoffs.  In some settings learning roles as part of the training process may be necessary or advantageous. If two independent agents have never interacted with one another before, they are unlikely to share a convention for negotiating roles.  In a fully cooperative task, hierarchical learning will be unnecessary, and it would be desirable if both players learned to use the follower's faster gradient update.

Our goal then is to allow each player to learn which role (leader or follower) will yield the highest payoffs given its partner's behavior.  A straightforward approach is to wrap the hierarchical learning process in a ``meta-learning'' process, where a pair of independent meta-learners each commits to a particular role for some pre-determined number of steps, and then evaluates their average payoff under that role given the role their partner chose.  Importantly, the meta-learning process is \textit{symmetric}, with no leader-follower hierarchy.

Algorithm~\ref{alg:meta_learning} describes the meta-learning update for player 1 (the update for player 2 only differs in the use of the payoff function $f_2$).  The length of the meta-learning intervals are described by a fixed schedule $\{ \tau_n \}_{n \geq 0}$, which, as with the Hi-C update, we assume will grow arbitrarily large over time.  This commitment schedule, along with the step-size schedule $\{ \beta_n \}_{n \geq 0}$, is assumed to be shared between both meta-learners.  Each meta-learner maintains a ``meta-strategy'' parameterized by a single scalar value $w_n \in \Re$.  We let $p(w_n) = \sin^2 (w_n)$ be the probability of choosing to lead at interval $n$, with $1 - p(w_n) = \cos^2 (w_n)$ being the probability of choosing to follow.  This parameterization allows $w$ to be unbounded, and allows us to represent \textit{pure} strategies using finite values of $w$.  At each interval $n$ the meta-learner for player 1 approximates gradient ascent on its expected payoff
\begin{equation}
    f_n(w^1) = [p(w^1), 1-p(w^1)] {\bf\hat{ U}}^{1}_n [p(w^{2}_n), 1-p (w^{2}_n)]^{\top}
\end{equation}
where $w^{2}_n$ denotes player 2's meta-strategy, while the matrix ${\bf\hat{ U}}^{1}_n$ denotes the expected average payoff for player $1$ under each of the four possible joint role assignments. Note that the expectation ${\bf\hat{ U}}^{1}_n$ is conditioned on the underlying joint strategy $(x, y)$ at the start of interval $n$.  As with the Hi-C update, the meta-learner updates $w$ using the single-sample variant of SPSA.  Note that other approaches to estimating the gradient $\nabla_{w^1} f_n(w^1)$, for example, using the log-likelihood trick, would suffer from singularities when evaluated at pure strategies (e.g., $w^{1} = 0$ or $w^{1} = \frac{\pi}{2}$). 

The fact that ${\bf\hat{U}}^{1}_n$ may depend on the underlying joint strategy means that, in general, we cannot apply standard stochastic approximation results to the meta-learning process alone.  If we can assume, however, that the long-term behavior of underlying learning process is asymptotically independent of its initial state (for any possible leader-follower role assignment) then we can describe the joint meta-learning process by the stochastic approximation
\begin{align}
    w^{1}_{n+1} \!\! &= w^{1}_n \!+\! \beta_n \left[[\nabla p(w^{1}_n), -\nabla p(w^{1}_n)] {\bf U}^{1} [p(w^{2}_n), 1-p(w^{2}_n)]^{\top} \right. \nonumber \\
    &\hspace{1.5cm}\left. + \:\zeta^{1}_n + \varepsilon^{1}_n \right] \\
    w^{2}_{n+1} \!\!&= w^{1}_n \!+\! \beta_n \left[ [p(w^{1}_n), 1-p(w^{1}_n)] {\bf U}^{2} [\nabla p(w^{2}_n), -\nabla p(w^{2}_n)]^{\top} \right. \nonumber \\
    &\hspace{1.5cm}\left. + \:\zeta^{2}_n + \varepsilon^{2}_n \right]
\end{align}
where $U^1 = \text{E}[\lim_{n \rightarrow \infty} \hat{U}^{1}_n]$ and $U^2 = \text{E}[\lim_{n \rightarrow \infty} \hat{U}^{2}_n]$, $\zeta^{1}_n$ and $\zeta^{2}_n$ are the noise terms introduced by SPSA, and $\epsilon^{1}_n$ and $\epsilon^{2}_n$ are $o(1)$.  This in turn converges (under the standard SA assumptions) to an ICT invariant set of the limiting ODE:
\begin{align}
    \label{eqn:meta_learning_limit}
    \dot{w}^{1} &= [\nabla p(w^{1}), -\nabla p(w^{1})] {\bf U}^{1} [p(w^{2}), 1-p(w^{2})]^{\top} \\
    \dot{w}^{2} &= [p(w^{1}), 1-p(w^{1})] {\bf U}^{2} [\nabla p(w^{2}), -\nabla p(w^{2})]^{\top}
\end{align}
Whether such an invariant set necessarily corresponds to a specific leader--follower ordering will depend on the structure of the game.  In general, such a set may not correspond to an equilibrium point of the ODE, with the players never converging to fixed roles.  We leave the characterization of games in which role negotiation can be guaranteed to converge as an open question for future work.

\section{Discussion}
\label{discussion}

A major motivation for our work is to understand the problem of ad hoc collaboration between autonomous agents, both AI and human. In this case, agents cannot assume anything about how others' behavior will change over time, and need to adapt to one another simultaneously. Previous analysis of \textit{naive} simultaneous learning updates such as SGA has suggested that such learning processes may be highly unstable, and may fail to converge to good joint strategies. Research in differentiable games has in recent years focused on the types of centralized training settings commonly arising in deep learning, where some learners must have detailed knowledge of other's loss functions and learning updates. Thus, these methods and their analyses are not directly applicable to ad hoc collaboration. Hierarchical learning dynamics are well-suited to this setting, but have previously required that the leader have direct access to the follower's payoff function. Our work overcomes this critical limitation. Our approach also has the potential to be useful in centralized training. Compared to the coupled hierarchical gradient update, \awnn will generally have much lower per-step computational cost, though whether this offsets the potential increase in sample complexity in practice is an open question.

\paragraph{Future work.} Immediate future directions for this work include expanding the class of follower learning updates and payoff functions for which we can provide concrete convergence guarantees. This could include more flexible methods such as stochastic gradient descent, or no-regret learning rules such as online mirror descent. Recent work on bi-level optimization such as \citet{liu2021towards} has also provided theoretical tools for analysing convergence in the case of non-concave follower objectives.  The extension of these results to our uncoupled setting is another important question for future work. 
%Additionally, empirical evaluation of how \awnn compares to the coupled hierarchical gradient update, or other coupled approaches such as LOLA when scaled to high-dimensional training problems such as GANs or multi-agent reinforcement learning is an important direction for future research. 
Finally, there are a number of open questions regarding the dynamics of role negotiation.  These include determining in which classes of games the players will converge to fixed roles with high probability, and whether the players' average payoffs can be guaranteed to converge even when the roles themselves do not.

% An important question for the hierarchical play is the leader--follower role assignment. Previous work so far assumed the roles are either preassigned, or can be dictated by one of the agents (see section \ref{related_work}). This is not suitable for settings with fully autonomous agents such as ad hoc collaboration. Whether agents can negotiate the role assignment online as part of the game dynamics was an open question since \citet{basar1973relative}. Our answer is a partial yes, and we present the first step towards embedding the online negotiation dynamics into the underlying game dynamics.

\section{Related Work}
\label{related_work}

\paragraph{Differentiable Games.} 
Previous work on gradient ascent in differentiable games has found that simultaneous gradient ascent on individual payoff functions can fail to converge~\cite{mescheder2017gans,mertikopoulos2018}. This has motivated the development of alternative solution concepts that are better suited to differentiable games, such as chain recurrent sets~\cite{papadimitriou2018recurrent} and local Stackelberg equilibria~\cite{jin2020local}. Others have proposed modified gradient ascent approaches to achieve at least local convergence to fixed-points in certain classes of games~\cite{mescheder2017gans, balduzzi2018mechanics, schafer2019competitive}. Similar to our approach are methods for two-player games that update the individual strategies on two different timescales~\cite{metz2017unrolled, nouiehed2019minmax, mazumdar2019nash}. As with our approach,~\citet{nouiehed2019minmax} implement timescale separation by having the follower execute multiple gradient steps for every leader update, though unlike our work, their leader does not directly attempt to \textit{shape} the behavior of the follower.

\paragraph{Hierarchical Model of Play and Role Assignment.} 
Assuming the follower plays an immediate best-response, previous work has also provided lower bounds on the sample complexity of identifying Stackelberg equilibria in Stackelberg security games~\cite{peng2019learning}, bandit games~\cite{bai2021sample} and Markov games~\cite{ramponi2022stackelberg}.  The challenge in our setting is that we must assume the follower is implementing an incremental learning update, which may only play a true best-response asymptotically.  Most closely related to our work is the two-timescale hierarchical gradient update~\cite{fiez2020implicit, zheng2022stackelberg}. Unlike our method, the hierarchical gradient update requires that the leader have access to the follower's payoffs.  The earliest analysis of leader--follower role assignments was in \citet{basar1973relative}, while \citet{basar1982feedback} considered the case where the roles switch between players depending on an exogenous process. More recent work has considered the case where players change roles depending on the game state, where the roles are still pre-assigned for each state \cite{bacsar2010differential}. In the context of strategic classification, \citet{NEURIPS2021_812214fb} have analysed the setting where a specific player can choose and dictate a role assignment for everyone. 
%Additionally, unlike our work, they also require that the leader's objective is convex, the follower is a no-regret learner, and the leader has a fixed update interval.
To our knowledge, ours is the first result on online negotiation of the roles during hierarchical play.

\paragraph{Multiagent Learning.} 
Our work is also related to \textit{opponent shaping} approaches~\cite{foerster2018lola,willi2022cola}, where one or both learners explicitly account for their partner's learning behavior, and update their strategy accordingly.  Of these the model-free opponent shaping (M-FOS) framework of~\citet{mfos} is closest to ours.  The key differences from our method are that M-FOS assumes the follower can be ``reset" after each interval, and only allows the follower to adapt for a fixed number of stages. In contrast, we do not require such resets, and explicitly account for the fact that the follower's strategy depends on the entire history of interaction.  \awnn also allows the follower to learn over increasing time horizons, enabling asymptotic convergence.  Finally, \awnn is conceptually similar to no-regret learning methods for non-stationary tasks~\cite{farias2003experts} and adaptive partners~\cite{poland2005universal}, in which the leader commits to candidate ``expert'' strategies for increasingly long time intervals.

\paragraph{Bi-level Optimization.}  The problem of finding differential Stackelberg equilibria can be cast as bi-level optimization.  Indeed, the hierarchical gradient update (\cite{fiez2020implicit, zheng2022stackelberg}) corresponds to an approximate implicit differentiation (AID) method for bi-level problems.  Iterative differentiation (ITD) methods (e.g. \cite{pedregosa2016hyperparameter,franceschi2018bilevel, ji2021bilevel,grazzi2020iteration}) are conceptually similar to our approach as well. However, both AID and ITD methods require analytically differentiating through the follower's best-response function, which in turn requires the gradients (and Hessians) of the follower's payoff function. Recent work~\citep{liu2022bome} does not use Hessians, but still requires knowledge of the follower's objective functions. Developed for centralized training settings such as GANs, these methods cannot be applied to settings where the learners are truly autonomous and decentralised.  While some recent work (\cite{chen2023bilevel,maheshwari2023convergent}) has presented zeroth-order (gradient-free) methods for bi-level optimization, these simulate multiple independent copies of the follower, and so require access to the follower's payoffs and learning update.

\section{Conclusion}
\label{conclusion}

We have presented, to the best of our knowledge, the first \textit{uncoupled} learning update that can be shown to converge to differential Stackelberg solutions for a broad class of general-sum differentiable games. The \awnn learning update for the leader agent can be implemented without access to the follower's payoff function or the details of their learning update. This also means that \awnn does not need to estimate the gradients or Hessians of the follower's payoffs.  Most importantly, our convergence results provide theoretical insights into uncoupled hierarchical learning processes, where one agent must learn about the preferences of another agent through its observable behavior alone. We have also presented the first online role negotiation dynamics, which illustrate how agents can strategically negotiate a leader--follower ordering as part of the hierarchical learning process.

%%%%%%%%%%%%%%%%%%%%%%%%%%%%%%%%%%%%%%%%%%%%%%%%%%%%%%%%%%%%%%%%%%%%%%%%

%%% The acknowledgments section is defined using the "acks" environment
%%% (rather than an unnumbered section). The use of this environment 
%%% ensures the proper identification of the section in the article 
%%% metadata as well as the consistent spelling of the heading.

\begin{acks}
This work was supported by the Academy of Finland (Flag-ship programme: Finnish Center for Artificial Intelligence, FCAI; grants 319264, 313195, 305780,
292334, 328400, 28400), the UKRI Turing AI World-Leading Researcher Fellowship EP/W002973/1, the
Finnish Science Foundation for Technology and Economics (KAUTE), and the Hybrid Intelligence Center, a 10-year programme funded by the Dutch Ministry of Education, Culture and Science through the Netherlands Organisation for Scientific Research, https://hybrid-intelligence-centre.nl.
\end{acks}

%%%%%%%%%%%%%%%%%%%%%%%%%%%%%%%%%%%%%%%%%%%%%%%%%%%%%%%%%%%%%%%%%%%%%%%%

%%% The next two lines define, first, the bibliography style to be 
%%% applied, and, second, the bibliography file to be used.

\bibliographystyle{ACM-Reference-Format} 
\bibliography{references}

%%%%%%%%%%%%%%%%%%%%%%%%%%%%%%%%%%%%%%%%%%%%%%%%%%%%%%%%%%%%%%%%%%%%%%%%

\end{document}